\documentclass[]{fairmeta}
\usepackage[algoruled]{algorithm2e}
\usepackage{amsthm,amsmath,amssymb}
\usepackage{graphicx}
\usepackage{natbib}
\usepackage[hidelinks,breaklinks,hypertexnames=False]{hyperref}
\usepackage{subfig}
\newtheorem{theorem}{Theorem}[]

\newtheorem{remark1}[theorem]{Remark}

\DeclareMathOperator{\E}{\mathop{}\mathbb{E}}
\DeclareMathOperator{\Id}{{\rm Id}}
\DeclareMathOperator{\var}{\mathop{}{Var}}
\renewcommand{\epsilon}{\varepsilon}

\title{Guarantees of confidentiality via Hammersley-Chapman-Robbins bounds}
\author[1]{Kamalika Chaudhuri}
\author[1]{Chuan Guo}
\author[1]{Laurens van der Maaten}
\author[1]{Saeed Mahloujifar}
\author[1]{Mark Tygert}
\affiliation[1]{Fundamental Artificial Intelligence Research at Meta}
\date{\today}
\correspondence{Mark Tygert at \email{mark@tygert.com}}

\begin{document}

\abstract{Protecting privacy during inference with deep neural networks
is possible by adding noise to the activations in the last layers prior
to the final classifiers or other task-specific layers.
The activations in such layers are known as ``features''
(or, less commonly, as ``embeddings'' or ``feature embeddings'').
The added noise helps prevent reconstruction of the inputs
from the noisy features.
Lower bounding the variance of every possible unbiased estimator
of the inputs quantifies the confidentiality arising from such added noise.
Convenient, computationally tractable bounds are available
from classic inequalities of Hammersley and of Chapman and Robbins
--- the HCR bounds. Numerical experiments indicate that the HCR bounds are
on the precipice of being effectual for small neural nets with the data sets,
``MNIST'' and ``CIFAR-10,'' which contain 10 classes each
for image classification. The HCR bounds appear to be insufficient on their own
to guarantee confidentiality of the inputs to inference
with standard deep neural nets, ``ResNet-18'' and ``Swin-T,''
pre-trained on the data set, ``ImageNet-1000,'' which contains 1000 classes.
Supplementing the addition of noise to features with other methods
for providing confidentiality may be warranted in the case of ImageNet.
In all cases, the results reported here limit consideration to amounts
of added noise that incur little degradation in the accuracy of classification
from the noisy features. Thus, the added noise enhances confidentiality
without much reduction in the accuracy on the task of image classification.}

\maketitle

\section{Introduction}

The Hammersley-Chapman-Robbins (HCR) bounds of~\cite{hammersley}
and~\cite{chapman-robbins} provide easily interpretable, tight data-driven
guarantees on confidentiality. The interpretation is especially simple
and direct, lower bounding the variance of any estimator
for reconstructing inputs to inference with neural networks.
Moreover, computing the bounds is efficient and straightforward.
Confidentiality stems from suitable addition of noise;
the HCR bounds quantify the effect of the added noise
on the minimum possible variance of any estimator.

This paper studies the privacy preservation arising from adding noise
to the activations in the final layers of deep neural networks,
that is, to the layers immediately preceding the classifiers used during
supervised training of the nets and used for classification during inference.
The most common terminology for these activations is ``features'' (or vectors
of features). Less common synonyms are ``feature embeddings''
or simply ``embeddings.'' Adding noise is also known as ``dithering.''
Dithering features is a canonical method for limiting the quality
of possible reconstructions of the inputs
that generated the noiseless features.

The present paper omits consideration of adding noise directly to the data
or for proxies to that process, such as DP-SGD of~\cite{dp-sgd},
since quantifying their privacy preservation in terms
of the variance minimized over all possible estimators is trivial,
given directly by the amount of noise added.

A method for quantifying privacy that is closely related to the HCR bounds
is to use Fisher information and the Cram\'er-Rao bound,
as advanced by~\cite{hannun-guo-van-der-maaten} and others.
The Cram\'er-Rao bound is most useful
when a quadratic form specified by the Fisher information matrix
is a good approximation to the expected loss (the risk function)
near the parameters at which the Fisher information is evaluated.
In contrast, the HCR bounds used below are typically tight
whether or not a quadratic form specified by Fisher information
evaluated at a single setting of parameters is a good approximation.
Furthermore, evaluating the Fisher information for complicated models
in machine learning such as deep neural networks can be difficult
or costly at scale. The approach proposed below avoids most of the difficulties
and is computationally tractable. In fact, the HCR bounds always exist, whereas
the Cram\'er-Rao bounds exist only when the loss is sufficiently smooth.

The present paper could be viewed as providing an alternative
to the Cram\'er-Rao bounds that is more practical.
Indeed, the Cram\'er-Rao bound is a certain limit of an HCR bound.
Subsection~\ref{cramer-rao} below details the connection,
a connection first made in the original works
of~\cite{hammersley} and of~\cite{chapman-robbins}.

Unfortunately, the experimental results reported below indicate that
the HCR bounds are weak for one of the data sets tested,
at least with some standard neural nets for image classification
(image classification is also known as ``image recognition'').
Section~\ref{conclusion} below concludes that dithering features
and leveraging the associated HCR bounds would be most useful
in conjunction with cruder, brute-force methods for enhancing confidentiality
of the inputs to inference (such as limiting the sizes
of the vectors of features being revealed).

Earlier work, notably the thesis of~\cite{alisic}, also applies HCR bounds
to quantify privacy, comparing favorably with the differential privacy
of~\cite{dwork-roth}.
The focus of~\cite{alisic} and the subset of the thesis reported
by~\cite{alisic-molinari-pare-sandberg} is on confidentiality in the measurement
of dynamical systems --- quite different from the setting considered
in the present paper --- yet the results are complementary and consonant
with those of the present paper.

The remainder of the present paper has the following structure:
Section~\ref{methods} develops theory and algorithms based on HCR bounds.
Section~\ref{results} reports numerical experiments with the popular data sets
MNIST, CIFAR-10, and ImageNet-1000 for image classification, when processed
with standard architectures such as ResNets and Swin Transformers
as well as with some especially simple,
illustrative neural nets.\footnote{Permissively licensed open-source codes that
can automatically reproduce all results of the present paper
are available at \url{https://github.com/facebookresearch/hcrbounds}}
Section~\ref{conclusion} draws several conclusions and suggests coupling
the methods presented in the present paper with cruder, brute-force techniques
for enhancing confidentiality.

\section{Methods}
\label{methods}

This section details the methodology of the present paper.
Subsection~\ref{hcrbounds} briefly reviews the general formulation
of Hammersley-Chapman-Robbins (HCR) bounds.
Subsection~\ref{outputs} specifies the HCR bounds for dithering vectors
of features specifically and develops algorithms for computing the bounds.
Subsection~\ref{additive} specializes the HCR bounds to the addition
(to the features) of independent and identically distributed noise.
Subsection~\ref{cramer-rao} considers a limit in which the HCR bounds
become the classical Cram\'er-Rao bounds (under suitable assumptions
of regularity in the parametric model, so that the relevant derivatives exist).

\subsection{Hammersley-Chapman-Robbins bounds}
\label{hcrbounds}

This subsection reviews a classic bound introduced independently
by~\cite{hammersley} and~\cite{chapman-robbins}.
Specifically, we use the multivariate generalization detailed, for example,
by~\cite{wikipedia}.

We consider a family of probability density functions (pdfs) $f_{\theta}(x)$
of a vector $x$, parameterized by a vector $\theta$,
with $x$ coming from an $n$-dimensional real vector space $\mathbb{R}^n$
and $\theta$ coming from a $p$-dimensional real vector space $\mathbb{R}^p$.
We consider any estimator $\hat\theta(X)$ of $\theta$;
the estimator is a function of the vector $X$ of observations.
We define $g(\theta)$ to be the expected value of $\hat\theta$
with respect to the pdf $f_{\theta}$, that is,
\begin{equation}
g(\theta) = \E_{\theta}[\hat\theta]
= \int_{\mathbb{R}^n} \hat\theta(x) \, f_{\theta}(x) \, dx.
\end{equation}

The Hammersley-Chapman-Robbins (HCR) bound is
\begin{equation}
\label{hcr}
\var_{\theta}(\hat\theta_k)
\ge \frac{(g_k(\theta + \epsilon) - g_k(\theta))^2}
         {\E_{\theta}\left[\left(
          \frac{f_{\theta + \epsilon}(X)}{f_{\theta}(X)} - 1\right)^2 \right]}
\end{equation}
for $k = 1$, $2$, \dots, $p$
and for any vector $\epsilon$ in the same $p$-dimensional real vector space
$\mathbb{R}^p$ to which $\theta$ belongs.
In~(\ref{hcr}),
$\hat\theta_k$ is the $k$th entry of the vector-valued $\hat\theta$,
similarly $g_k$ is the $k$th entry of the vector-valued $g$, and
\begin{equation}
\var_{\theta}(\hat\theta_k)
= \E_{\theta}\left[ (\hat\theta_k - g_k(\theta))^2 \right]
= \int_{\mathbb{R}^n} \left( \hat\theta_k(x)
- \int_{\mathbb{R}^n} \hat\theta_k(y) \, f_{\theta}(y) \, dy \right)^2
\, f_{\theta}(x) \, dx
\end{equation}
and
\begin{equation}
\E_{\theta}\left[\left(
\frac{f_{\theta + \epsilon}(X)}{f_{\theta}(X)} - 1\right)^2 \right]
= \int_{\mathbb{R}^n}
\left(\frac{f_{\theta + \epsilon}(x)}{f_{\theta}(x)} - 1\right)^2
\, f_{\theta}(x) \, dx.
\end{equation}

If $\hat\theta$ is an unbiased estimator of $\theta$, then $g(\theta) = \theta$
and~(\ref{hcr}) simplifies to
\begin{equation}
\label{hcru}
\var_{\theta}(\hat\theta_k)
\ge \frac{(\epsilon_k)^2}{\E_{\theta}\left[\left(
          \frac{f_{\theta + \epsilon}(X)}{f_{\theta}(X)} - 1\right)^2 \right]}
\end{equation}
for $k = 1$, $2$, \dots, $p$
and for any vector $\epsilon$ in the same $p$-dimensional real vector space
$\mathbb{R}^p$ to which $\theta$ belongs.
Requiring the estimator to be unbiased is tantamount to forbidding
the use of extra, outside information such as a Bayesian prior.
Unbiasedness is a reasonable yet significant restriction.
If, for example, the actual values of the data are known from sources
other than the observations $X$, then clearly the estimator can be better
than unbiased --- the estimator could simply ignore the observations $X$
and report the correct values known a-priori from another source.

A bound on the mean-square error of any unbiased estimator $\hat{\theta}$
of $\theta$ follows immediately from~(\ref{hcru}):
\begin{equation}
\label{mse}
\E_{\theta}\left[ \frac{1}{p} \sum_{k=1}^p (\hat\theta_k - \theta_k)^2 \right]
\ge \frac{\frac{1}{p} \sum_{k=1}^p (\epsilon_k)^2}{\E_{\theta}\left[\left(
          \frac{f_{\theta + \epsilon}(X)}{f_{\theta}(X)} - 1\right)^2 \right]}
\end{equation}
for any vector $\epsilon$ in the same $p$-dimensional real vector space
$\mathbb{R}^p$ to which $\theta$ belongs.

\subsection{Dithering the features of a machine-learned model}
\label{outputs}

This subsection discusses how to enhance privacy
(specifically, confidentiality) of the input data used
during inference with an already trained machine-learned model,
by adding noise to the features that the inference calculates.
The formal term for adding noise is ``dithering.''
The present subsection specializes the HCR bounds of~(\ref{hcr})
and~(\ref{hcru}) to this setting and details algorithms
for computing the bounds.

To set notation, we let the vector $\theta$ of parameters denote the input data
and the vector $X$ of observations denote the resulting features,
with noise added to the features.
We let the vector $a_{\theta}$ of activations denote the features
without noise added.
Note that the input data need not be the entire test set, but
could be only one or more of the individual examples input during inference.

Obtaining a tight HCR bound hinges on selecting a suitable vector $\epsilon$
of perturbations, perhaps taking the maximum bound realized
over several choices of $\epsilon$.
The ideal $\epsilon$ maximizes the ratio in the HCR bound of~(\ref{hcru}).
When the noise added to the features is Gaussian,
with the entries of the noise being independent and identically distributed
centered normal variates, the denominator in~(\ref{hcru})
becomes the expression in~(\ref{normalcase}) given below.
Maximizing~(\ref{hcru}) thus amounts to making the perturbation $\epsilon$
to the input $\theta$ as large as possible while making
the corresponding perturbation $z_{\epsilon}$
to the features $a_{\theta}$ as small as possible,
for $z_{\epsilon}$ of~(\ref{activepert}) below;
that is, the goal is to maximize the ratio of Euclidean norms
$\|\epsilon\| / \|z_{\epsilon}\|$, or, equivalently,
to minimize the ratio $\|z_{\epsilon}\|/ \|\epsilon\|$.
If the perturbation $\epsilon$ is small, then linearization yields that
$z_{\epsilon} \approx (\partial a_{\theta} / \partial \theta) \, \epsilon$,
which is the product of the Jacobian $(\partial a_{\theta} / \partial \theta)$
and the perturbation $\epsilon$ to $\theta$.
Under this linear approximation, the minimum of the ratio
$\|z_{\epsilon}\| / \|\epsilon\|$ is therefore equal to reciprocal
of the spectral norm of the pseudoinverse
of the Jacobian $(\partial a_{\theta} / \partial \theta)$;
after all, the spectral norm of the pseudoinverse is simply the reciprocal
of the least singular-value of the Jacobian itself,
and the least singular-value of the Jacobian
$(\partial a_{\theta} / \partial \theta)$
is by definition the minimum of the ratio
$\|(\partial a_{\theta} / \partial \theta) \, \epsilon\| / \|\epsilon\|
\approx \|z_{\epsilon}\| / \|\epsilon\|$.

Some simple iterations can approximate the spectral norm
of the pseudoinverse of the Jacobian $(\partial a_{\theta} / \partial \theta)$
while simultaneously calculating a perturbation $\epsilon$ for which the ratio
$\|(\partial a_{\theta} / \partial \theta) \, \epsilon\| / \|\epsilon\|
\approx \|z_{\epsilon}\| / \|\epsilon\|$ is nearly minimal.
Indeed, iterations of LSQR of~\cite{paige-saunders} with the Jacobian
$(\partial a_{\theta} / \partial \theta)$ applied to vectors generated
during the iterations of LSQR and with the transpose
$(\partial a_{\theta} / \partial \theta)^\top$ of the Jacobian
applied to other vectors generated during the iterations
can approximate the action of the pseudoinverse of the Jacobian
$(\partial a_{\theta} / \partial \theta)$.
Such iterations of LSQR produce a perturbation $\epsilon$,
from which the corresponding perturbation $z_{\epsilon}$ to the features
is straightforward to calculate.
Then the newly calculated $z_{\epsilon}$ can serve as the starting point
$(\partial a_{\theta} / \partial \theta)^\top z_{\epsilon}$
in the normal equations for further iterations of LSQR.
The further iterations of LSQR yield an updated perturbation $\epsilon$,
from which the corresponding perturbation $z_{\epsilon}$ to the features
is straightforward to compute. Repeating this process several
(say, $i = 10$) times, iteratively updating $\epsilon$ and $z_{\epsilon}$
every time, will approximately minimize the ratio
$\|z_{\epsilon}\| / \|\epsilon\|$.
Algorithm~\ref{perturbation} provides pseudocode summarizing the procedure.

Note that automatic differentiation can apply to arbitrary vectors
both the Jacobian and its transpose, efficiently and matrix-free
(never actually having to form the full Jacobian).
Furthermore, there is no need to compute $\epsilon$ especially precisely,
as any approximation whatsoever to the ideal for $\epsilon$ yields
a perfectly rigorous guarantee via the HCR bounds.
Indeed, given the perturbation $\epsilon$ to the input $\theta$,
calculating the corresponding perturbation $z_{\epsilon}$ to the features
exactly (without any approximations) requires just one forward run
of inference with the machine-learned model.

\LinesNumbered
\begin{algorithm}[t]
\caption{Calculation of a perturbation $\epsilon$ to the vector of parameters
$\theta$}
\label{perturbation}
\DontPrintSemicolon
\KwIn{Positive integers $i$, $n$, and $p$,
a vector $z$ whose $n$ entries are real numbers,
a vector $\theta$ whose $p$~entries are real numbers, and functions $t$ and $u$
that apply the transpose of the Jacobian
$(\partial a_{\theta} / \partial \theta)$ and the Jacobian itself
(without transposition) to arbitrary vectors, respectively,
where $a_{\theta}$ is the vector of features
introduced in Subsection~{\ref{outputs}}; here,
$i$ is the number of repetitions of the LSQR algorithm of~\cite{paige-saunders}
that the present Algorithm~\ref{perturbation} will execute,
$z$ is the starting vector for the iterations of LSQR
(so $t(z)$ is the starting vector with regard to the normal equations),
and $\theta$ is the unperturbed input data.}
\KwOut{A vector $\epsilon$ whose $p$ entries are real-valued
and a vector $z_{\epsilon}$ whose $n$ entries are real-valued;
$\epsilon$ is the perturbation to $\theta$ such that
$z_{\epsilon} = a_{\theta + \epsilon} - a_{\theta}$.}

Set $z^{(0)} = z$.

Calculate the vector of features $a_{\theta}$ corresponding
to the input $\theta$.

\For{j {\rm = 1}, {\rm 2}, \dots, $i$}{

Set $\tilde{z}^{(j-1)} = \|z\| \cdot z^{(j-1)} / \|z^{(j-1)}\|$,
so that the Euclidean norms of $z$ and $\tilde{z}^{(j-1)}$ are equal.

Solve the least-squares problem of minimizing the Euclidean norm
$\| (\partial a_{\theta} / \partial \theta) \, \epsilon^{(j)}
- \tilde{z}^{(j-1)} \|$,
obtaining the minimizing $\epsilon^{(j)}$ using LSQR of~\cite{paige-saunders}.
LSQR should invoke the functions $t$ and $u$ to perform
the matrix-vector multiplications that LSQR requires.
This step 5 amounts to applying the pseudoinverse of the Jacobian
$(\partial a_{\theta} / \partial \theta)$ to $\tilde{z}^{(j-1)}$,
yielding $\epsilon^{(j)}$.

Calculate the vector of features $a_{\theta + \epsilon^{(j)}}$ corresponding
to the perturbed input $(\theta + \epsilon^{(j)})$.

Set $z^{(j)} = a_{\theta + \epsilon^{(j)}} - a_{\theta}$.

}

\Return{\rm $\epsilon = \epsilon^{(i)}$ and $z_{\epsilon} = z^{(i)}$.}

\end{algorithm}

\subsection{Additive noise}
\label{additive}

This subsection specializes the HCR bounds of the previous subsections
to the case in which the dithered features are simply the features
plus independent and identically distributed noise.
In particular, this subsection derives an explicit expression
for the right-hand side of~(\ref{hcru}).
In accordance with the notation of the preceding subsection,
Subsection~\ref{outputs},
we denote by $a_{\theta}$ the features generated by inference
with the already trained model using the original data $\theta$
(or just a single test example) as inputs,
and we denote by $a_{\theta + \epsilon}$ the features generated by inference
using the perturbed data $(\theta + \epsilon)$ as inputs.

Dithering yields the observed noisy vector of features
\begin{equation}
X = a_{\theta} + Z,
\end{equation}
where $Z$ is the noise added.
Denoting by $f$ the probability density function of the noise, we see that
\begin{equation}
\label{unperturbed}
f_{\theta}(X) = f_{\theta}(a_{\theta} + Z) = f(Z)
\end{equation}
and
\begin{equation}
\label{perturbed}
f_{\theta + \epsilon}(X) = f_{\theta}(X - (a_{\theta + \epsilon} - a_{\theta}))
= f_{\theta}(a_{\theta} + Z - (a_{\theta + \epsilon} - a_{\theta}))
= f(Z - (a_{\theta + \epsilon} - a_{\theta}))
= f(Z - z_{\epsilon}),
\end{equation}
where
\begin{equation}
\label{activepert}
z_{\epsilon} = a_{\theta + \epsilon} - a_{\theta};
\end{equation}
that is, $z_{\epsilon}$ is the perturbation added to the features
during determination of $\epsilon$ (with $z_{\epsilon}$ updated
to correspond exactly to the $\epsilon$ actually used,
as discussed at the end of Subsection~\ref{outputs}).
Combining~(\ref{unperturbed}) and~(\ref{perturbed}) yields that
the denominator in~(\ref{hcru}) is
\begin{equation}
\label{denom}
\E_{\theta}\left[\left(
           \frac{f_{\theta + \epsilon}(X)}{f_{\theta}(X)} - 1\right)^2 \right]
= \E\left[\left( \frac{f(Z-z_{\epsilon})}{f(Z)} - 1\right)^2 \right],
\end{equation}
where $z_{\epsilon}$ from~(\ref{activepert}) is viewed
as a fixed constant during evaluation of the expectation.

For some distributions of $Z$ ---
including the multivariate normal distribution $N(0, \sigma^2 \cdot \Id)$
corresponding to a standard deviation $\sigma$ ---
we can evaluate~(\ref{denom}) via analytic integration,
aligning one of the axes of integration
in the integral corresponding to the right-hand side of~(\ref{denom})
with the fixed direction given by $z_{\epsilon}$.
Appendix~\ref{normalcalc} performs the calculation for this normal case,
yielding that the denominator in~(\ref{hcru}) is
\begin{equation}
\label{normalcase}
\E_{\theta}\left[\left(
           \frac{f_{\theta + \epsilon}(X)}{f_{\theta}(X)} - 1\right)^2 \right]
= \exp\left( \frac{\|z_{\epsilon}\|^2}{\sigma^2} \right) - 1,
\end{equation}
where $\|z_{\epsilon}\|$ is the Euclidean norm of $z_{\epsilon}$
from~(\ref{activepert}). For more complicated distributions,
we can estimate the right-hand side of~(\ref{denom}) via Monte-Carlo methods.
For isotropic (that is, rotation-invariant) distributions
of the added noise $Z$,
such as the multivariate normal distribution $N(0, \sigma^2 \cdot \Id)$,
the value of~(\ref{denom}) depends only on the Euclidean norm
of $z_{\epsilon}$ and not on the entries of $z_{\epsilon}$ individually.
In all cases, the value of~(\ref{denom}) is independent
of the machine-learned model used.

\subsection{Cram\'er-Rao bounds}
\label{cramer-rao}

This subsection connects the earlier subsections with the famous approach
of Cram\'er and Rao --- a connection that the original works
of~\cite{hammersley} and of~\cite{chapman-robbins} note
as motivation for developing their own bounds.
(The remainder of this subsection will be assuming tacitly,
without further elaboration, that all derivatives required
for this subsection's derivations actually exist and are continuous.
Unlike the HCR bounds, the Cram\'er-Rao bounds pertain only to scenarios
in which the derivatives do exist.)

If the perturbation $\epsilon$ is very small,
then $z_{\epsilon} = a_{\theta + \epsilon} - a_{\theta}$
will also be very small, with $z_0 = 0$, so
\begin{equation}
\label{local1}
(z_{\epsilon})_j
= \sum_{k=1}^p \frac{\partial (z_{\epsilon})_j}{\partial \epsilon_k}
  \, \epsilon_k + o(\|\epsilon\|)
\end{equation}
for $j = 1$, $2$, \dots, $n$, while the right-hand side of~(\ref{normalcase})
becomes
\begin{equation}
\label{local2}
\exp\left( \frac{\|z_{\epsilon}\|^2}{\sigma^2} \right) - 1
= \frac{\|z_{\epsilon}\|^2}{\sigma^2} + o(\|\epsilon\|^2)
= \frac{1}{\sigma^2} \sum_{j=1}^n ((z_{\epsilon})_j)^2 + o(\|\epsilon\|^2),
\end{equation}
where $(z_{\epsilon})_j$ denotes the $j$th entry of the vector $z_{\epsilon}$.
Combining~(\ref{local1}) and~(\ref{local2}) yields
\begin{equation}
\label{anyeps}
\exp\left( \frac{\|z_{\epsilon}\|^2}{\sigma^2} \right) - 1
= \frac{1}{\sigma^2} \sum_{j=1}^n \left( \sum_{k=1}^p
  \frac{\partial (z_{\epsilon})_j}{\partial \epsilon_k} \, \epsilon_k \right)^2
  + o(\|\epsilon\|^2).
\end{equation}

Evaluating~(\ref{anyeps}) for a perturbation $\epsilon$ in which all entries
but one --- say the $k$th --- are zero yields
\begin{equation}
\label{oneeps}
\exp\left( \frac{\|z_{\epsilon}\|^2}{\sigma^2} \right) - 1
= \frac{(\epsilon_k)^2}{\sigma^2} \sum_{j=1}^n
  \left( \frac{\partial (z_{\epsilon})_j}{\partial \epsilon_k} \right)^2
  + o(\|\epsilon\|^2),
\end{equation}
where $k$ is one of the positive integers $1$, $2$, \dots, $p$. Naturally,
\begin{equation}
\label{inverse}
\frac{\partial (z_{\epsilon})_j}{\partial \epsilon_k}
= \frac{1}{\partial \epsilon_k/\partial (z_{\epsilon})_j}
\end{equation}
for $j = 1$, $2$, \dots, $n$.
Combining~(\ref{hcru}), (\ref{normalcase}), and~(\ref{oneeps})
and taking the limit $\epsilon \to 0$ then yields
\begin{equation}
\label{hcrugaussian}
\var_{\theta}(\hat\theta_k)
\ge \frac{\sigma^2}
         {\sum_{j=1}^n (\partial (z_{\epsilon})_j/\partial \epsilon_k)^2}
= \frac{\sigma^2}
       {\sum_{j=1}^n 1/(\partial \epsilon_k/\partial (z_{\epsilon})_j)^2}
\end{equation}
for $k = 1$, $2$, \dots, $p$,
where the latter equality in~(\ref{hcrugaussian}) follows from~(\ref{inverse}).
Please note that~(\ref{hcrugaussian}) is exact, not approximate ---
the higher-order terms vanish in the limit $\epsilon \to 0$.
Evaluating the bound~(\ref{hcrugaussian}) for all $p$ values of $k$
requires the computation of either $p$ or $n$ gradients,
where $p$ is the dimension of the space of parameters
and $n$ is the dimension of the space of observations.
(Taking the Jacobian of $z_{\epsilon}$ with respect to $\epsilon$
requires $n$ gradients; taking the Jacobian of $\epsilon$
with respect to $z_{\epsilon}$ requires $p$ gradients.)
The inequality in~(\ref{hcrugaussian}) is known as the ``Cram\'er-Rao bound,''
as elucidated by~\cite{hannun-guo-van-der-maaten}, for example.

\section{Results}
\label{results}

This section applies the methods of the previous section,
Section~\ref{methods}, to several standard data sets and neural architectures
for classifying the input images.\footnote{Permissively licensed open-source
software that can automatically reproduce all the results reported here
is available at \url{https://github.com/facebookresearch/hcrbounds}}
All bounds reported in the present section are for the standard deviations
corresponding to~(\ref{hcru}); of course, the standard deviation
is the square root of the variance from~(\ref{hcru}).
Subsection~\ref{mnist} considers MNIST,
a classic data set of $28 \times 28$ pixel grayscale scans
of handwritten digits, first training a simple neural net
on the standard training set and then conducting inference
and computing the associated HCR bounds on the test set.
Subsection~\ref{cifar10} does similarly for CIFAR-10, a classic data set
of $32 \times 32$ pixel color images of 10 classes, namely
airplanes, birds, boats, cars, cats, deer, dogs, frogs, horses, and trucks.
Subsection~\ref{imagenet} considers ImageNet, a standard data set
with 1000 classes, processing images from the validation set
via the conventional pre-trained neural nets, ``ResNet18'' and ``Swin-T,''
from TorchVision of~\cite{torchvision}.

In the coming subsections, ``Affine$_{m \times n}$'' refers to a layer
which multiplies the input row vector from the right
by an $m \times n$ matrix whose entries are learned and adds a vector which is
independent of the input (that is, a ``bias'') that is also learned;
the dimension of the input is $m$ and the dimension of the output is $n$.
``ReLU'' refers to a layer which preserves unchanged every non-negative entry
of the input and zeros every negative entry; the dimensions of the input
and of the output are the same.
``Flatten'' refers to a layer which reshapes the input into a single,
longer vector.
``Convolution2D$_{m \times n{\rm (channels)}; k \times \ell{\rm (kernel)}}$''
refers to a layer which convolves each of the $m$ channels of the input
with $n$ convolutional kernels, each of size $k \times \ell$ pixels
whose values are learned, and adds to the result an image which is
independent of the input (that is, a ``bias'') that is also learned.
``MaximumPooling2D$_{m \times n{\rm (stride)}; k \times \ell{\rm (kernel)}}$''
refers to a layer which partitions the input into $m \times n$ blocks of pixels
and replaces each block with the maximum value in the block
(in this paper, the stride and size of the kernel are always the same,
that is, $m = k$ and $n = \ell$);
the first dimension of the output is $1/m$ times the first dimension
of the input, while the second dimension of the output is $1/n$ times
the second dimension of the input.
``Softmax'' refers to a layer which calculates the softmax of the input vector
(the softmax is also known as the ``Gibbs distribution'');
the dimensions of the input and of the output are the same.

The weights and biases in the neural networks are the learned values;
the values of the inputs, features, and class-confidences are activations
(that is, values at nodes) in the neural nets. All results reported
are HCR bounds maximized over 25 independent and identically distributed
pseudorandom realizations of $z_{\epsilon}$ in~(\ref{normalcase}),
obtained by running the algorithm (Algorithm~\ref{perturbation})
of Subsection~\ref{outputs} with the $n$ entries of the starting vector $z$
being proportional to the normally distributed noise added to the features.
The constant of proportionality is $1/\sqrt{n}$ times the size $s$
of perturbation specified in the captions to the subfigures
(these sizes are $1/200$, $1/500$, and $1/1000$ for the different subfigures,
as indicated in the captions).
This constant of proportionality results in the right-hand side
of~(\ref{normalcase}) being roughly $\exp(s^2) - 1 \approx s^2$,
where $s$ is the size of the perturbation
($s = 1/200$, $s = 1/500$, or $s = 1/1000$, as specified
in the subfigures' captions).

\subsection{MNIST}
\label{mnist}

This subsection reports the results of numerical experiments
with the standard data set, ``MNIST,''
a data set presented by~\cite{lecun-bottou-bengio-haffner}.
MNIST contains images of handwritten digits (0, 1, 2, \dots, 9).

To calculate features for a given input, we use the activations
in the last layer of the following neural network:
inputs $\rightarrow$
Flatten $\rightarrow$
Affine$_{784 \times 784}$ $\rightarrow$
ReLU $\rightarrow$
Affine$_{784 \times 784}$ $\rightarrow$
ReLU $\rightarrow$
features

There are 784 entries both in the input vector for each image
and in the corresponding features. The input images are $28 \times 28$ pixels,
with only a single color channel (the inputs are grayscale).

Given the features, the classifier takes the argmax of the activations
(values at the nodes) in the last layer of the following neural network,
passing the given features as inputs to the network:
features $\rightarrow$
Affine$_{784 \times 10}$ $\rightarrow$
Softmax $\rightarrow$
class-confidences

In all processing, we first normalize the pixels' potential values
to range from 0 to 1, then subtract the overall mean 0.1037, and finally divide
by the standard deviation 0.3081. When displaying images,
we reverse all these normalizations.

For training, we use random minibatches of 32 examples each,
over 6 epochs of optimization (thus sweeping 6 times
through all 60,000 examples from the training set of MNIST).
We minimize the empirical average cross-entropy loss
using AdamW of~\cite{loshchilov-hutter}, with a learning rate of 0.001.

On the test set of MNIST, the average accuracy for classification
without dithering is 97.9\% and with dithering is 95.1\%.

\begin{figure}
\begin{center}
\subfloat[10 iterations with a perturbation of 1/200]{
\includegraphics[width=0.4\textwidth]
{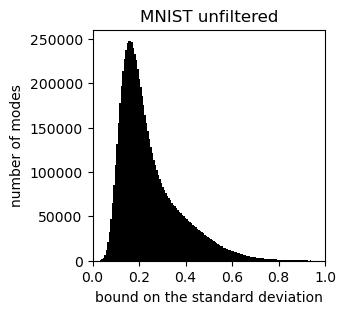}
}
\subfloat[10 iterations with a perturbation of 1/200]{
\includegraphics[width=0.4\textwidth]
{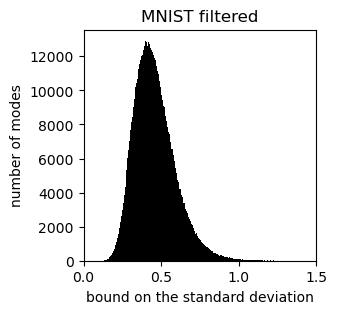}
}

\subfloat[10 iterations with a perturbation of 1/1000]{
\includegraphics[width=0.4\textwidth]
{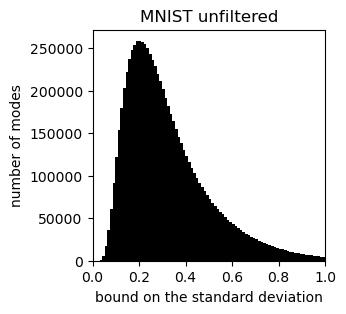}
}
\subfloat[10 iterations with a perturbation of 1/1000]{
\includegraphics[width=0.4\textwidth]
{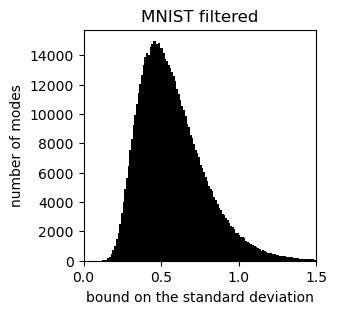}
}
\end{center}
\caption{Histograms of the HCR bounds over the 10,000 examples
of MNIST's test set,
both unfiltered and filtered to the $8 \times 8$ lowest-frequency modes
of the type-2 discrete cosine transform; the numbers of iterations
are the numbers of repetitions of LSQR in Subsection~\ref{outputs},
which is the input $i$ in Algorithm~\ref{perturbation}}
\label{mnisth}
\end{figure}

\begin{figure}
\begin{center}
\subfloat[one, perturbed by 1/1000]{
\includegraphics[width=0.3\textwidth]
{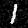}
}
\subfloat[one, perturbed by 1/200]{
\includegraphics[width=0.3\textwidth]
{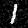}
}
\subfloat[one, unperturbed]{
\includegraphics[width=0.3\textwidth]
{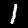}
}

\subfloat[four, perturbed by 1/1000]{
\includegraphics[width=0.3\textwidth]
{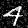}
}
\subfloat[four, perturbed by 1/200]{
\includegraphics[width=0.3\textwidth]
{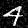}
}
\subfloat[four, unperturbed]{
\includegraphics[width=0.3\textwidth]
{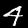}
}

\subfloat[nine, perturbed by 1/1000]{
\includegraphics[width=0.3\textwidth]
{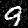}
}
\subfloat[nine, perturbed by 1/200]{
\includegraphics[width=0.3\textwidth]
{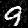}
}
\subfloat[nine, unperturbed]{
\includegraphics[width=0.3\textwidth]
{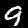}
}
\end{center}
\caption{Reconstructions of examples from MNIST's test set}
\label{mnisti}
\end{figure}

In Figures~\ref{mnisth} and~\ref{mnisti}, the size of the perturbation
(either 1/200 or 1/1000) pertains to the Euclidean norm of $z_{\epsilon}$
in~(\ref{normalcase}). In the limit that the size is 0, the HCR bounds
would become Cram\'er-Rao bounds (if the parameterizations
of the neural networks were differentiable), as in~(\ref{hcrugaussian}).
The results for the different sizes turn out to be reasonably similar.

Figure~\ref{mnisth} histograms (over all examples in the test set)
the magnitudes of the HCR lower bounds on the standard deviations
of unbiased estimators for the original images' values.
The estimates are for the Fourier modes in a discrete cosine transform (DCT)
of type~II, with the DCT normalized to be an orthogonal linear transformation
(meaning real and unitary or isometric).
The modes of the DCT form an orthonormal basis suitable as a system
of coordinates; note that these modes are for the normalized input images,
standardized such that the standard deviation of the normalized pixel values
is 1 and the mean is 0. The histograms in the rightmost column
of Figure~\ref{mnisth} consider only the $8 \times 8$ lowest-frequency modes,
whereas the histograms in the leftmost column consider all $28 \times 28$.

Figure~\ref{mnisth} shows that the bounds would have been reasonably effective
had the pixels of the original images not been mostly almost pure black
or pure white (so that rounding away the obtained bounds denoises the estimates
very effectively).

Figure~\ref{mnisti} visualizes the HCR bounds on three examples
from the test set. The visualization involves
(1)~adding to the modes of the DCT for the normalized original image
the product of independent and identically distributed Rademacher variates
(which are $-1$ with probability 1/2 and $+1$ with probability 1/2)
times the corresponding HCR bounds, (2) inverting the DCT,
and (3) reversing the per-pixel normalization back into the conventional
perceptual space in which the values of pixels can range from 0 to 1
(while clipping negative values to 0 and clipping values exceeding 1 to 1).
Figure~\ref{mnisti} illustrates that the obtained bounds are significant
yet ineffective (mostly since thresholding the grayscale images
to purely black-and-white would denoise away much
of the displayed perturbations).

\subsection{CIFAR-10}
\label{cifar10}

This subsection presents the results of numerical experiments
with the standard benchmark data set, ``CIFAR-10,'' of~\cite{krizhevsky}.
CIFAR-10 contains images representing ten labeled classes
--- airplanes, birds, boats, cars, cats, deer, dogs, frogs, horses, and trucks.

To calculate features for a given input, we use the activations
in the last layer of the following neural network,
adapted from the net of~\cite{shahrestani}:
inputs $\rightarrow$
Convolution2D$_{\rm 3 \times 32(channels); 3 \times 3(kernel)}$ $\rightarrow$
ReLU $\rightarrow$
MaximumPooling2D$_{\rm 2 \times 2(stride); 2 \times 2(kernel)}$ $\rightarrow$
Convolution2D$_{\rm 32 \times 1024(channels); 5 \times 5(kernel)}$ $\rightarrow$
ReLU $\rightarrow$
MaximumPooling2D$_{\rm 3 \times 3(stride); 3 \times 3(kernel)}$ $\rightarrow$
Convolution2D$_{\rm 1024 \times 3072(channels); 3 \times 3(kernel)}$
$\rightarrow$
ReLU $\rightarrow$
Flatten $\rightarrow$
Affine$_{3072 \times 3072}$ $\rightarrow$
ReLU $\rightarrow$
features
(``Flatten'' simply removes dimensions that are 1 in the shape of the tensor,
in this particular case.)

There are 3,072 entries both in the input vector for each image
and in the corresponding features. The input images are $32 \times 32$ pixels,
with three color channels (red, green, and blue).

Given the features, the classifier takes the argmax of the activations
(values at the nodes) in the last layer of the following neural network,
passing the given features as inputs to the network:
features $\rightarrow$
Affine$_{3072 \times 10}$ $\rightarrow$
Softmax $\rightarrow$
class-confidences

In all processing, we first normalize the pixels' potential values
to range from 0 to 2 and then subtract 1, so that the resulting pixel values
can range from $-1$ to 1. When displaying images,
we reverse all these normalizations.

For training, we use random minibatches of 32 examples each,
over 7 epochs of optimization (thus sweeping 7 times
through all 50,000 examples from the training set of CIFAR-10).
We use the AdamW optimizer of~\cite{loshchilov-hutter} with a learning rate
of 0.001, minimizing the empirical average cross-entropy loss.

On 2,500 examples drawn at random without replacement
from the test set of CIFAR-10, the average accuracy for classification
without dithering is 70\% and with dithering is 50\%.

\begin{figure}
\begin{center}
\subfloat[6 iterations with a perturbation of 1/500]{
\includegraphics[width=0.4\textwidth]
{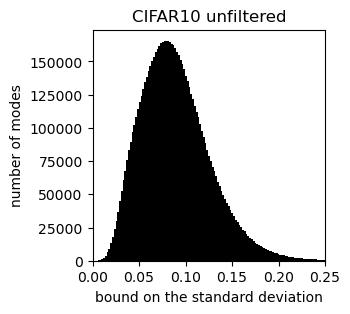}
}
\subfloat[6 iterations with a perturbation of 1/500]{
\includegraphics[width=0.4\textwidth]
{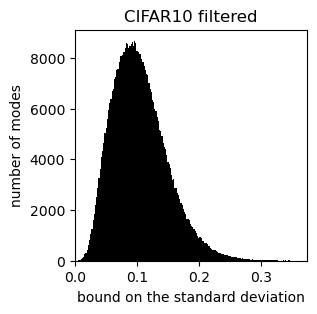}
}

\subfloat[4 iterations with a perturbation of 1/1000]{
\includegraphics[width=0.4\textwidth]
{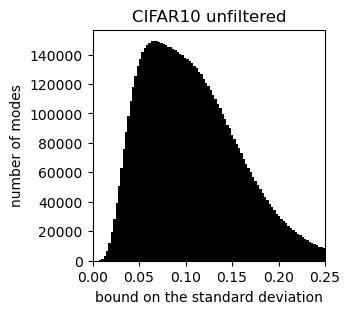}
}
\subfloat[4 iterations with a perturbation of 1/1000]{
\includegraphics[width=0.4\textwidth]
{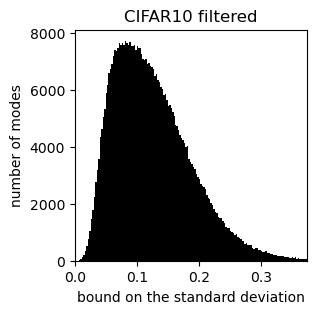}
}
\end{center}
\caption{Histograms of the HCR bounds over 2,500 examples
from CIFAR-10's test set,
both unfiltered and filtered to the $8 \times 8$ lowest-frequency modes
of the type-2 discrete cosine transform; the numbers of iterations
are the numbers of repetitions of LSQR in Subsection~\ref{outputs},
which is the input $i$ in Algorithm~\ref{perturbation}}
\label{cifar10h}
\end{figure}

\begin{figure}
\begin{center}
\subfloat[airplane, perturbed by 1/1000]{
\includegraphics[width=0.3\textwidth]
{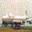}
}
\subfloat[airplane, perturbed by 1/500]{
\includegraphics[width=0.3\textwidth]
{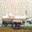}
}
\subfloat[airplane, unperturbed]{
\includegraphics[width=0.3\textwidth]
{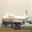}
}

\subfloat[bird, perturbed by 1/1000]{
\includegraphics[width=0.3\textwidth]
{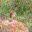}
}
\subfloat[bird, perturbed by 1/500]{
\includegraphics[width=0.3\textwidth]
{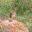}
}
\subfloat[bird, unperturbed]{
\includegraphics[width=0.3\textwidth]
{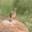}
}

\subfloat[boat, perturbed by 1/1000]{
\includegraphics[width=0.3\textwidth]
{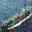}
}
\subfloat[boat, perturbed by 1/500]{
\includegraphics[width=0.3\textwidth]
{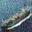}
}
\subfloat[boat, unperturbed]{
\includegraphics[width=0.3\textwidth]
{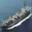}
}
\end{center}
\caption{Reconstructions of examples from CIFAR-10's test set}
\label{cifar10i}
\end{figure}

In Figures~\ref{cifar10h} and~\ref{cifar10i}, the size of the perturbation
(either 1/500 or 1/1000) pertains to the Euclidean norm of $z_{\epsilon}$
in~(\ref{normalcase}). The size 1/1000 is close to the limit
in which HCR bounds would become Cram\'er-Rao bounds (if the parameterizations
of the neural networks were differentiable), as in~(\ref{hcrugaussian}).
The results for the different sizes are quite similar.

Figure~\ref{cifar10h} histograms (over 2,500 examples from the test set)
the magnitudes of the HCR lower bounds on the standard deviations
of unbiased estimators for the original images' values.
The estimates are for the Fourier modes in a discrete cosine transform (DCT)
of type~II, with the DCT normalized to be orthogonal (meaning real and unitary
or isometric).
The modes of the DCT form an orthonormal system of coordinates;
note that these modes are for the normalized input images,
standardized such that the normalized pixel values range from $-1$ to $1$.
The histograms in the rightmost column of Figure~\ref{cifar10h}
consider only the $8 \times 8$ lowest-frequency modes,
while the histograms in the leftmost column consider all $32 \times 32$.

Figure~\ref{cifar10i} visualizes the HCR bounds on three examples
from the test set. As with Figure~\ref{mnisti}, the visualization involves
(1) adding to the values of the modes in the DCT
for the normalized original image the product
of independent and identically distributed Rademacher variates
with the corresponding HCR bounds on the standard deviations,
(2) inverting the DCT, and
(3) reversing the per-pixel normalization back to where the values of pixels
in each color channel can range from 0 to 1 (while clipping negative values
to 0 and clipping values exceeding 1 to 1).

Both Figure~\ref{cifar10h} and Figure~\ref{cifar10i} show that the bounds
are on the precipice of guaranteeing that decent reconstructions
of the original images are impossible from the dithered features.

\subsection{ImageNet-1000}
\label{imagenet}

This subsection presents the results of numerical experiments
with the popular data set, ``ImageNet-1000,'' of~\cite{imagenet}.
ImageNet-1000 contains a thousand labeled classes, each consisting
of images representing a particular noun (such as a species or a dog breed).

All examples of the present subsection consider 128 examples
from the validation set of ImageNet-1000, drawing the examples uniformly
at random without replacement. These 128 examples are more than sufficient
to find that the HCR bounds for ImageNet are ineffective,
allowing the dithered features to lead to full reconstructions
that are imperceptibly different from the input images. 
All models of the present subsection are trained
on the training set of ImageNet;
we downloaded the pre-trained networks from PyTorch's ``model zoo''
of~\cite{torchvision}.
The input images get resized to be $91 \times 91$ pixels
(with three color channels --- red, green, and blue)
and then upsampled to be $224 \times 224$ pixels
(with the same RGB color channels) for input to the pre-trained neural nets.
There are slightly fewer degrees of freedom in an image
that has $91 \times 91$ pixels for each of three color channels
than in the features of either of the pre-trained networks
(``ResNet-18'' and ``Swin-T'') considered here.
For pre-processing, we applied to the input images the usual normalizations
from the model zoo of~\cite{torchvision}.

In Figures~\ref{resnet18h} and~\ref{swinth}, the size (either 1/500 or 1/1000)
of the perturbation pertains to the Euclidean norm of $z_{\epsilon}$
in~(\ref{normalcase}).
The size 1/1000 is close to 0 --- close to the limit in which HCR bounds
would become Cram\'er-Rao bounds (if the parameterizations
of the neural networks were differentiable), as in~(\ref{hcrugaussian}).
The results of the different sizes are similar.

The following two subsubsections refrain from displaying
analogues of Figures~\ref{mnisti} and~\ref{cifar10i},
since visualizations of which reconstructions are possible (analogous to those
of Figures~\ref{mnisti} and~\ref{cifar10i}) turn out to be perceptually
indistinguishable from the original images.

\subsubsection{ResNet-18}
\label{resnet-18}

This subsubsection uses the ResNet-18 of~\cite{he-zhang-ren-sun}.
There are 24,843 entries in the input vector for each image
and 25,088 entries in the corresponding features.
The average (top-1) accuracy of classification without dithering is 57\%
and with dithering is 54\%.

\begin{figure}
\begin{center}
\subfloat[10 iterations with a perturbation of 1/500]{
\includegraphics[width=0.4\textwidth]
{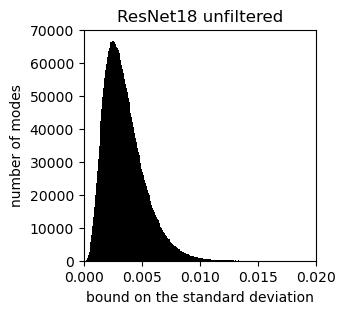}
}
\subfloat[10 iterations with a perturbation of 1/500]{
\includegraphics[width=0.4\textwidth]
{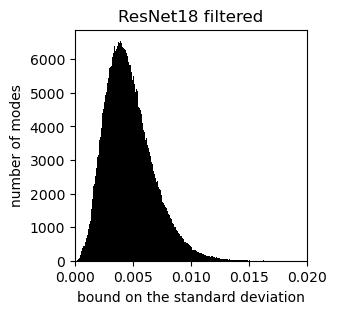}
}

\subfloat[10 iterations with a perturbation of 1/1000]{
\includegraphics[width=0.4\textwidth]
{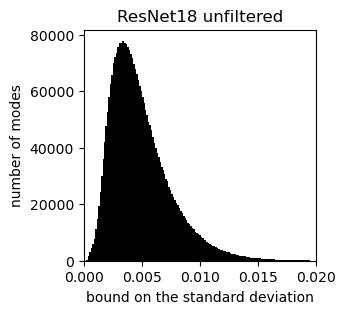}
}
\subfloat[10 iterations with a perturbation of 1/1000]{
\includegraphics[width=0.4\textwidth]
{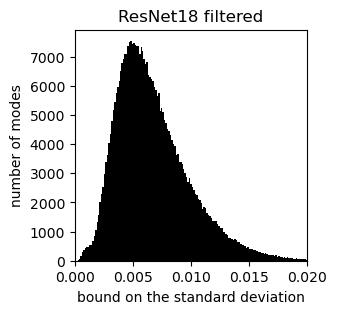}
}
\end{center}
\caption{Histograms of the HCR bounds over 128 examples
from ImageNet's validation set, using a ResNet-18,
both unfiltered and filtered to the $32 \times 32$ lowest-frequency modes
of the type-2 discrete cosine transform; the numbers of iterations
are the numbers of repetitions of LSQR in Subsection~\ref{outputs},
which is the input $i$ in Algorithm~\ref{perturbation}}
\label{resnet18h}
\end{figure}

Figure~\ref{resnet18h} histograms (over 128 examples from the validation set)
the magnitudes of the HCR lower bounds on the standard deviations
of unbiased estimators for the original images' values.
The estimates are for the Fourier modes in an orthogonal
discrete cosine transform (DCT) of type~II.
The modes of the DCT form an orthonormal system of coordinates;
note that these modes are for the normalized input images,
standardized such that the standard deviation of the normalized pixel values
is about 1 and the mean is roughly 0.
The rightmost histograms in Figure~\ref{resnet18h}
consider only the $32 \times 32$ lowest-frequency DCT modes.

The bounds reported in Figure~\ref{resnet18h} are useless
for all practical purposes, providing next to no guarantee of any protection
against reconstruction attacks.

\subsubsection{Swin-T}
\label{swin}

This subsubsection uses the Swin-T of~\cite{liu-lin-cao-hu-wei-zhang-lin-guo}.
There are 24,843 entries in the input vector for each image
and 37,632 entries in the corresponding features.
The average (top-1) accuracy of classification without dithering is 64\%
and with dithering is 54\%.

\begin{figure}
\begin{center}
\subfloat[10 iterations with a perturbation of 1/500]{
\includegraphics[width=0.4\textwidth]
{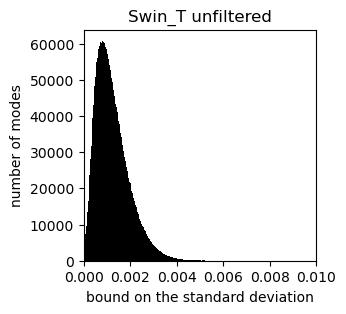}
}
\subfloat[10 iterations with a perturbation of 1/500]{
\includegraphics[width=0.4\textwidth]
{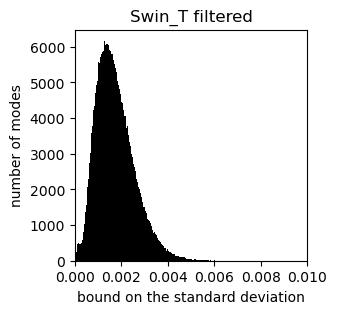}
}

\subfloat[10 iterations with a perturbation of 1/1000]{
\includegraphics[width=0.4\textwidth]
{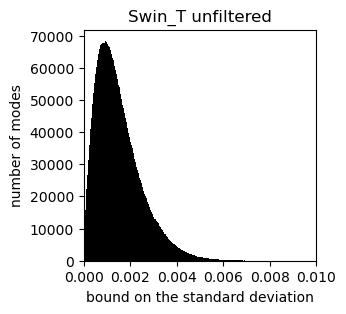}
}
\subfloat[10 iterations with a perturbation of 1/1000]{
\includegraphics[width=0.4\textwidth]
{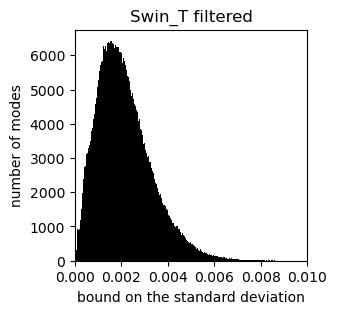}
}
\end{center}
\caption{Histograms of the HCR bounds over 128 examples
from ImageNet's validation set, using a Swin-T,
both unfiltered and filtered to the $32 \times 32$ lowest-frequency modes
of the type-2 discrete cosine transform; the numbers of iterations
are the numbers of repetitions of LSQR in Subsection~\ref{outputs},
which is the input $i$ in Algorithm~\ref{perturbation}}
\label{swinth}
\end{figure}

Figure~\ref{swinth} histograms (over 128 examples) the magnitudes
of the HCR lower bounds on the standard deviations of unbiased estimators
for the original images' values. The estimates are for the modes
in an orthogonal discrete cosine transform of type~II.
The modes of the DCT constitute an orthonormal basis appropriate for a system
of coordinates; note that these modes are for the normalized input images,
standardized such that the standard deviation of the normalized pixel values
is around 1 and the mean is approximately 0.
The rightmost histograms in Figure~\ref{swinth}
filter down to the $32 \times 32$ lowest-frequency modes.

As with Figure~\ref{resnet18h}, the bounds reported in Figure~\ref{swinth}
provide effectively no guarantee of protection against reconstructing
the input images.

\section{Conclusion}
\label{conclusion}

The guarantees provided by the Hammersley-Chapman-Robbins (HCR) bounds
in the results presented above are sometimes on the precipice
of being very useful, but are far from ideal.
The results above consider only examples
in which the neural networks are at least somewhat deep.
The HCR bounds might be more useful a-priori for shallow neural-networks
such as those corresponding to popular generalized linear models.
However, for such models Cram\'er-Rao bounds are easy to calculate
and simpler than the HCR analogues; none of the computational sophistication
developed in the present paper is necessary
to compute ideal Cram\'er-Rao bounds in such cases.
\cite{hannun-guo-van-der-maaten} took this approach.

Thus, the HCR approach appears to be ineffectual on its own.
Perhaps the best use of the HCR bounds would be to supplement
other, cruder techniques for enhancing privacy.
An obvious such cruder technique would be to limit the sizes
of the vectors of features. In the examples considered above,
the sizes of the vectors of features are never less than the corresponding
numbers of pixels in the images times the numbers of color channels.
In the complete absence of noise, reconstructing the whole original images
from the calculated features can be possible only when the number of features
is no less than the number of pixel values being reconstructed
(though, even then, computational cost might limit the feasibility
of full reconstruction in practice).
In the presence of noise, the HCR bounds rigorously limit the quality
of the reconstruction. Yet the above results indicate that the bounds
are fairly ineffectual for large models.
A more effective strategy than relying exclusively on the HCR bounds
could be to limit the sizes of the vectors of features.
After all, the numbers of degrees of freedom in the original images
that any scheme whatsoever can reconstruct from the corresponding features
obviously cannot ever be greater than the sizes of the vectors of features.
Dithering and the HCR bounds can nicely complement the limiting of the sizes
of the vectors of features.

\section*{Acknowledgements}

We would like to thank Awni Hannun and Edward Suh.

\appendix
\section{Normally distributed noise}
\label{normalcalc}

This appendix considers the multivariate normal distribution
in which all $n$ entries of a vector $Z$ are
independent and identically distributed as $N(0, \sigma^2)$,
so that the probability density function (pdf) of $Z$ is
\begin{equation}
\label{multivariaten}
f(z) = \frac{1}{(2 \pi \sigma^2)^{n/2}}
       \exp\left(-\frac{\|z\|^2}{2\sigma^2}\right),
\end{equation}
where $\|z\|$ is the Euclidean norm of $z$.
With this pdf, the right-hand side of~(\ref{denom}) is
\begin{equation}
\label{rotinv}
\E\left[\left( \frac{f(Z-z_{\epsilon})}{f(Z)} - 1\right)^2 \right]
= \int_{\mathbb{R}^n} \left( \frac{f(z - v)}{f(z)} - 1 \right)^2 f(z) \, dz,
\end{equation}
where $v$'s first entry $v_1 = \|z_{\epsilon}\|$ is the Euclidean norm
of $z_{\epsilon}$ and $v$'s other entries $v_k = 0$ for $k > 1$;
the invariance of~(\ref{multivariaten}) to rotations of the coordinate system
yields~(\ref{rotinv}) --- the right-hand side of~(\ref{rotinv})
aligns the first coordinate axis with the direction of $z_{\epsilon}$.
The remainder of this appendix simplifies~(\ref{rotinv}) further.

Substituting~(\ref{multivariaten}) into the right-hand side of~(\ref{rotinv})
yields
\begin{multline}
\label{combo}
\int_{\mathbb{R}^n} \left( \frac{f(z - v)}{f(z)} - 1 \right)^2 f(z) \, dz \\
= \frac{1}{(2 \pi \sigma^2)^{n/2}}
\int_{-\infty}^{\infty} \cdots \int_{-\infty}^{\infty}
\left( \exp\left(\frac{(z_1)^2 - (z_1-v_1)^2}{2 \sigma^2}\right) - 1 \right)^2
\exp\left(-\frac{(z_1)^2 + \dots + (z_n)^2}{2 \sigma^2}\right)
\, dz_1 \, dz_2 \cdots dz_n \\
= \frac{1}{(2 \pi)^{n/2}}
\int_{-\infty}^{\infty} \cdots \int_{-\infty}^{\infty}
\left( \exp\left(\frac{(z_1)^2 - (z_1-c)^2}{2}\right) - 1 \right)^2
\exp\left(-\frac{(z_1)^2 + \dots + (z_n)^2}{2}\right)
\, dz_1 \, dz_2 \cdots dz_n,
\end{multline}
where
\begin{equation}
c = \frac{v_1}{\sigma} = \frac{\|z_{\epsilon}\|}{\sigma}.
\end{equation}
Expanding the square yields three terms
\begin{equation}
\label{expansion}
\left( \exp\left(\frac{(z_1)^2 - (z_1-c)^2}{2}\right) - 1 \right)^2
= \exp\left((z_1)^2 - (z_1-c)^2\right)
- 2 \exp\left(\frac{(z_1)^2 - (z_1-c)^2}{2}\right) + 1.
\end{equation}
The last term in~(\ref{expansion}) corresponds in~(\ref{combo}) to
\begin{equation}
\frac{1}{(2 \pi)^{n/2}} \int_{-\infty}^{\infty} \cdots \int_{-\infty}^{\infty}
\exp\left(-\frac{(z_1)^2 + \dots + (z_n)^2}{2}\right)
\, dz_1 \, dz_2 \cdots dz_n = 1.
\end{equation}
The penultimate term in~(\ref{expansion}) corresponds in~(\ref{combo}) to
\begin{multline}
-\frac{2}{(2 \pi)^{n/2}} \int_{-\infty}^{\infty} \cdots \int_{-\infty}^{\infty}
\exp\left(-\frac{(z_1-c)^2 + (z_2)^2 + \dots + (z_n)^2}{2}\right)
\, dz_1 \, dz_2 \cdots dz_n \\
= -\frac{2}{(2 \pi)^{n/2}}
\int_{-\infty}^{\infty} \cdots \int_{-\infty}^{\infty}
\exp\left(-\frac{(z_1)^2 + (z_2)^2 + \dots + (z_n)^2}{2}\right)
\, dz_1 \, dz_2 \cdots dz_n = -2.
\end{multline}
The first term in the right-hand side of~(\ref{expansion})
corresponds in~(\ref{combo}) to
\begin{multline}
\frac{1}{(2 \pi)^{n/2}} \int_{-\infty}^{\infty} \cdots \int_{-\infty}^{\infty}
\exp\left(-\frac{2(z_1-c)^2 - (z_1)^2 + (z_2)^2 + \dots + (z_n)^2}{2}\right)
\, dz_1 \, dz_2 \cdots dz_n \\
= \frac{1}{\sqrt{2 \pi}} \int_{-\infty}^{\infty}
\exp\left(-\frac{2(z_1-c)^2 - (z_1)^2}{2}\right) \, dz_1
= \frac{1}{\sqrt{2 \pi}} \int_{-\infty}^{\infty}
\exp\left(-\frac{(z_1)^2 - 4cz_1 + 2c^2}{2}\right) \, dz_1.
\end{multline}
Further simplification yields
\begin{equation}
\frac{1}{\sqrt{2 \pi}} \int_{-\infty}^{\infty}
\exp\left(-\frac{(z_1)^2 - 4cz_1 + 2c^2}{2}\right) \, dz_1
= \frac{\exp(c^2)}{\sqrt{2 \pi}} \int_{-\infty}^{\infty}
\exp\left(-\frac{(z_1 - 2c)^2}{2}\right) \, dz_1
= \exp(c^2).
\end{equation}

Combining all formulas in this appendix yields that the right-hand side
of~(\ref{denom}) is
\begin{equation}
\E\left[\left( \frac{f(Z-z_{\epsilon})}{f(Z)} - 1\right)^2 \right]
= \exp\left(\frac{\|z_{\epsilon}\|^2}{\sigma^2}\right) - 1.
\end{equation}

\bibliography{paper}
\bibliographystyle{authordate1}

\end{document}